\def\year{2022}\relax
\documentclass[letterpaper]{article} 
\usepackage{aaai22}  
\usepackage{times}  
\usepackage{helvet}  
\usepackage{courier}  
\usepackage[hyphens]{url}  
\usepackage{graphicx} 
\usepackage{natbib}  
\usepackage{caption} 
\DeclareCaptionStyle{ruled}{labelfont=normalfont,labelsep=colon,strut=off} 
\usepackage{algorithm}
\usepackage{algorithmic}

\usepackage{newfloat}
\usepackage{listings}
\floatstyle{ruled}
\newfloat{listing}{tb}{lst}{}
\floatname{listing}{Listing}
\title{Symbols as a Lingua Franca for Bridging Human-AI Chasm\\ for Explainable and Advisable AI Systems\thanks{To appear in AAAI-22 Blue Sky track}}
\author{
Subbarao Kambhampati,  Sarath Sreedharan, Mudit Verma,  Yantian Zha, Lin Guan \\
{\tt(rao,ssreedh3,mverma13,yzha3,lguan9)@asu.edu}
\\
}
\affiliations{\textbf{School of Computing \& AI\\  
        Arizona State University}
}

\begin{document}

\maketitle

\begin{abstract}
Despite the surprising power of many modern AI systems that often learn their own representations, there is significant discontent about their inscrutability and the attendant problems in their ability to interact with humans. 
While alternatives such as  {\em neuro-symbolic} approaches have been proposed, there is a lack of consensus on what they are about. There are often two independent motivations (i) symbols as a {\em lingua franca} for human-AI interaction and (ii) symbols as system-produced abstractions used by the AI system in its internal reasoning. The jury is still out on whether AI systems will need to use symbols in their internal reasoning to achieve general intelligence capabilities. Whatever the answer there is, the need for (human-understandable) symbols in human-AI interaction seems quite compelling. Symbols, like emotions, may well not be {\em sine qua non} for intelligence {\em per se}, but they will be crucial for AI systems to interact with us humans -- as we can neither turn off our emotions nor get by without our symbols.
In particular, in many human-designed domains, humans would be interested in providing explicit (symbolic) knowledge and advice -- and expect machine explanations in kind. This alone requires AI systems to to maintain a {\em symbolic interface} for interaction with humans. 
In this \textit{blue sky} paper, we argue this point of view, and discuss research directions that need to be pursued to allow for this type of human-AI interaction.

\end{abstract}
\section{Introduction}

AI research community is grappling with an ongoing tussle between symbolic and non-symbolic approaches -- with the former using representations (and to some extent, knowledge) designed by the humans being often outperformed by the latter that  {\em learn their own representations}, but at the expense of inscrutability to humans in the loop. While \textit{neuro-symbolic systems} have received attention in some quarters \cite{lamb-neuro-symbolic,deraedt-neuro-symbolic},  the jury is still out on whether or not AI systems need internal symbolic reasoning to reach human-level intelligence. There are however compelling reasons for AI systems to communicate (take advice or provide explanations) from humans in essentially symbolic terms. After all, the alternatives would be either for the humans to understand the internal (learned) representations of the AI systems -- which seems like a rather poor way for us to design \textit{our} future; or for both humans and AI systems to essentially depend on the lowest common substrate they can exchange raw data -- be they images, videos or general \textit{space time signal tubes} (heretofore referred to as {\sl STST}). 

While STSTs -- be they saliency regions over images, or motion trajectories -- have been used in the machine learning community as a means to either advice or interpret the operation of AI systems \cite{greydanus2018visualizing,zhang2020atari,christiano-preferences}, 
 we contend that they will not scale to human-AI interaction in more complex sequential decision settings involving both tacit and explicit task knowledge \cite{polanyi-cacm}. This is because {\em exchanging information solely via STSTs presents intolerably high cognitive load for humans} -- which is what perhaps lead humans to evolve a symbolic language in the first place.\footnote{The urge to use symbolic representations for information exchange seems so strong that humans develop symbolic terms for speaking about concepts in even primarily tacit tasks (e.g. {\em pitch and roll} in basketball).} 

In this paper, we argue that orthogonal to the issue of whether AI systems use symbolic representations to support their internal reasoning, AI systems need to develop local symbolic representations that are interpretable to humans in the loop, and use them to take advice and/or give explanations for their decisions. The underlying motivations here are that human-AI interaction should be structured {\em for the benefit of the humans} -- thus the communication should be in terms that make most sense to humans. This argues for the inclusion of a symbolic interface,\footnote{The oft-repeated ``System 1/System 2" architectural separation \cite{kahneman2011thinking}, on the other hand,  doesn't strictly necessitate or lead to symbols that will serve as \textit{lingua franca}. If System 1/2 leads to symbols, they'll likely be as abstractions to improve efficiency. 
There is little reason to expect that abstractions that a pure learning system creates on its own, much like Wittgenstein's Lion, will wind up aligning well with the ones we humans use.}
especially in terms of symbols that already have meaning to the humans in the loop {(that is, these cannot just be internal symbolic abstractions that the machine may have developed for its own computational efficiency)}. 

Our argument is not that human-AI interaction must be exclusively in symbolic means -- but that it is crucial to also support a symbolic interface to make AI systems explainable and advisable.  {As we have previously argued \cite{polanyi-cacm}}, the inability of the current AI systems to take explicit knowledge-based advice, or provide interpretable explanations are at the root of many of the ills of the modern AI systems that learn their own internal representations.

Supporting such a \textit{lingua franca} symbolic interface brings up several significant challenges that need to be addressed by the research community: (1) the challenge of approximating the explanations -- and  constraining the interpretation of the human advice -- in terms of the symbolic interface, (2) the challenge of assembling the symbolic interface itself -- which in turn includes (2.1) getting the symbolic vocabulary and (2.2) grounding it in the representations that the AI system uses and (3) figuring out when and how to expand a preexisting symbolic vocabulary to improve the accuracy of  advice/explanation being communicated. In the remainder of the paper, we will discuss these challenges. We will, in addition, describe some of our recent approaches to tackle these challenges \cite{sreedharan2020bridging,guan2021widening,yantian-self-expl}. 


\section{Setup \& Uses of the Symbolic Interface}
\begin{figure}
\centering
\includegraphics[scale=0.25]{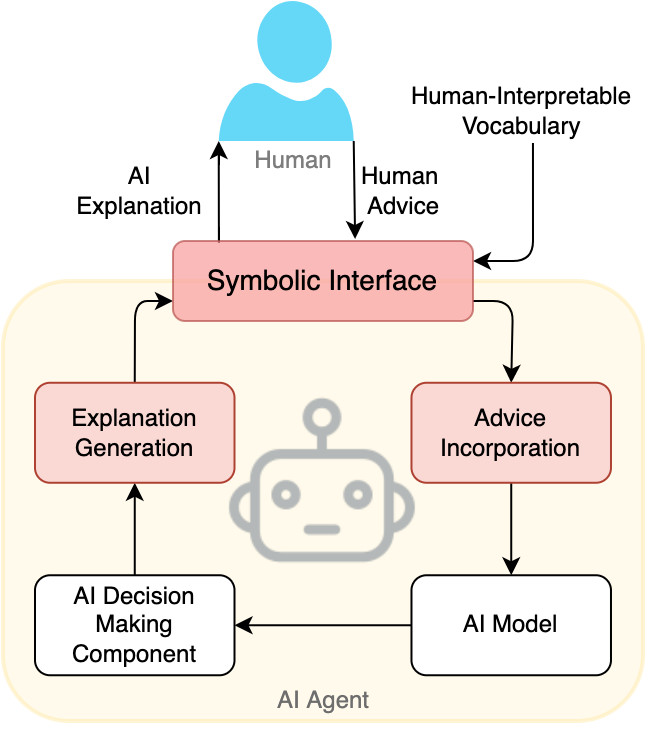}
\caption{Overall architecture of an AI system exposing a symbolic interface to a human user, 
enabling the AI agent to provide explanations to its decisions as well as accept guidance/preferences from the human in the form of advice.
}
\label{pipline}
\end{figure}
Figure \ref{pipline} presents the overview of an AI system capable of leveraging a symbolic interface of the type we advocate for in this paper. 
Note that in the architecture, the agent's decision-making (the white boxes) relies completely on its internal models, which may well be expressed, and be operating on, representations that are not directly accessible to the human; 
be they neural network based or based on some other internal symbolic abstraction.
While the agent could interact with the human across multiple modalities (such as annotated images, videos, demonstrations, etc.),
we are only arguing for the \textit{inclusion}, in addition, of a symbolic interface.
We specifically focus on symbolic interface expressed in human-understandable concepts.  The AI system uses the symbolic interface to both communicate its explanations to the user and to receive instructions and advice from them (the colored boxes in Figure~\ref{pipline}). In this section, we will discuss in more detail how this symbolic interface could be used towards these ends. 


\label{sec:post_hoc}


\medskip
\noindent 
{\bf Mental Models \& Symbolic Interface:} The ultimate use of the symbolic interface is that it allows the AI system to maintain and reason about human mental models \cite{rao-aimag-haai,yochan-xai-book}, in order to facilitate naturalistic human-AI interaction. These include
the human model $\mathcal{M}^H$ which captures human capabilities, preferences, and objectives; the agent model $\mathcal{M}^R$ which drives the agent decision-making, and the pivotal bridge model $\mathcal{M}^R_h$ that captures human expectations of the agent. 
It is by using $\mathcal{M}^R_h$ that the human decides what instructions and advice to give to the agent so they generate the desired outcomes, and it is by updating $\mathcal{M}^R_h$ that the robot could try to help the human make better sense of its behavior.
Naturally, $\mathcal{M}^R_h$ would need to reflect the human's understanding of the tasks, and as such in terms of the symbolic vocabulary constituting our lingua franca.

The foregoing implies that for the agent to influence the human's perception of it, and in turn to allow humans to provide additional information, it would need to build parallel symbolic representations of its own models as needed, and use them to reconcile human mental models.
These models should again be constructed using the elements from the shared symbolic vocabulary (see Section~\ref{sec-blackbox}) and may need to incorporate additional representational considerations like allowing for causal relations (such as structural causal models {\em a la} Pearl \citeyear{pearl2009causality}) and being compatible with folk psychological models of action and change (such as STRIPS/PDDL \cite{mcdermott1998pddl}).

\medskip
\noindent 
{\bf Explanations via Symbolic Interface:}
Prior work in explainable AI systems has already noted that explanations cannot be a \textit{soliloquy}, and must be in terms of the vocabulary and mental models of the recipients \cite{explain}. 
With the symbolic scaffolding in place,
we can now bring to bear all the previous developments in human-aware decision-making that build on symbolic models.
These include the explanation works (cf. \cite{yochan-xai-book},\cite{sreedharan2021foundations},\cite{sreedharan2021using}), using which the agent can  effectively make sense of its own decisions to the human in terms of these {\em post hoc}  models.
We shall see (Section~\ref{sec-vocab-expansion} and Section~\ref{sec-blackbox}) that the symbolic interface generalizes the ``explanation as model reconciliation" view adopted in these previous works, to include the challenge of reconciling differing vocabularies.



\medskip
\noindent
{\bf Advice via Symbolic Interface:} AI systems capable of taking high level advice from humans have been part of the holy grail of AI research ever since McCarthy's vision of Advice Taker \cite{mccarthy1959programs}.  The symbolic interface is a step towards realizing this vision, allowing humans to provide updates/advice for the AI system model in terms convenient to the humans.
For example, the human could specify additional constraints, previously unspecified preferences, causal dependencies \cite{pearl2009causality} etc. 
This includes works like reward-machines \cite{icarte2018using}  or restraining bolts \cite{de2019foundations}, wherein the advice-giver is effectively updating the AI behavior by introducing these new trajectory preferences. Note that we are talking about  {\em symbolic} advice over trajectories, rather than direct  trajectory level comparison (such as has been advocated by works such as \cite{christiano-preferences}). While communicating preferences by comparing STSTs (which is essentially what trajectory level comparisons do) is, in theory, feasible for preferences over both explicit and tacit tasks, it tends to be wildly cumbersome for explicit ones, and this is where symbolic interfaces will help make the communication more effective (see the discussion of EXPAND system in Section~\ref{expand-sec}).
{Additionally, in the case of providing human advice, by the virtue of the models built over symbolic interface being capable of capturing human expectations,  it can be used as a tool to better contextualize, build on, and generalize human input}. By leveraging the intuition that while coming up with the advice the human would have used a representation similar to the current interface, the AI system can potentially identify what objectives they may have had when providing the specific inputs (see the discussion of SERLfD in Section~\ref{expand-sec}).

\section{Research Challenges}

In this section, we will enumerate some of the open research challenges we need to address towards creating {and leveraging symbolic interfaces}.

\subsection{Collecting Initial Concept Set}
The first challenge is to collect the propositional and relational concepts that will form the basis of the interactions (and potentially even action labels). These symbols are meant to form the conceptual representation of the task and/or AI system's capabilities from the perspective of the human.  
Such symbols will be used to express any information the system may provide to the user, and to analyze any input the user may provide.
Note that in the case of user input, even when the input is not expressed in symbolic terms, the concepts provided allow the system to better utilize it. {For example, 
the agent may be provided a 
demonstration by the user, but access to state factors that may be important to the user will allow the agent to better generalize the demonstration \cite{yantian-self-expl}.}
These symbols may be obtained either directly or indirectly from the human. All the explicit knowledge representations -- including knowledge graphs -- will be in terms of symbols specified by a human. 
{In a similar vein, automatic extraction of these symbols should also be feasible, provided the symbols are taken from human vocabulary, as in the case of scene graph analysis \cite{krishna2017visual}}.
{We should note here that the previous works  that try to learn symbols from the perspective of the system (cf. \cite{konidaris2018skills}, \cite{bonet2019learning}, \cite{ghorbani2019towards}) need not result in useful symbolic interfaces, as those learned symbols may not make sense to humans.}

In general, we will assume there exists a way to map the STSTs to the corresponding set of symbols. One way to accomplish this may be to learn specific classifiers that identify the presence or absence of individual concepts of interest in the given slice of STST. 
As mentioned earlier, we could also use methods like scene-graph analysis that leverage the ability to identify common objects and their relationship to create a high-level symbolic representation of the relevant scene.
For everyday scenarios, which do not require specialized vocabulary, such methods can be particularly powerful, as this allows us to use robust systems to generate symbolic representations without the additional overhead of collecting domain specific-data. We are thus effectively amortizing the cost of collecting the concepts over the life-time of all AI systems that use them to generate the symbolic interface.
In the case where we are learning domain-specific vocabulary,  the concept set itself could come from multiple sources, including the user of the system and the system developers.
Even in this case, we could try to amortize the concept collection cost by creating domain-specific concept databases, which could be used by multiple systems (and for multiple users). Potential concept lists could also be mined from documents related to the domains.

\subsection{Learning Concept Grounding}
Once we have the initial concept set, the next challenge would be how to learn the  mappings between STST and symbolic concepts. The important point to recognize is that these groundings should try to approximate how the end-user would ground and understand the given concepts. In the case of off-the-shelf concept detection methods (like those that generate scene-graphs), the grounding is learned through large amounts of annotated data collected usually from crowd-sourced workers. In everyday scenarios, this is completely sufficient, as in general people tend to agree on the use of everyday concepts/words, etc (with possibly some cultural variations). On the other hand for more specialized domains, we may have to engage in a separate data-collection process to identify the grounding. In cases where such mappings are captured through learned classifiers, this may require us to collect positive and negative examples from the concept specifier. We also note that in general, the symbols in the interface will be grounded into a combination of the STSTs and the system's own learned internal representations. 
Note that any learned concept grounding is going to be noisy at best. This means that even if the concept classifiers or the scene-graph generator say certain concepts are present in a state or a slice of STST, it may not necessarily be true.
This means that any methods that are leveraging these symbols for downstream task should either be designed to handle the noise or allow for additional interaction with the end-user to resolve any mismatch in symbol grounding.

\subsection {Vocabulary Expansion}
\label{sec-vocab-expansion}
Another challenge that  AI systems would have to deal with in the long term is the fact that the original concept list would, in most cases, be incomplete. 
So the algorithms that work with these concept lists will need to explicitly allow for the fact that they may only have a portion of the total vocabulary list that the human may have access to. 
This means that in the process of interpreting or reasoning about the human input, or in building a symbolic approximation of its own model, the AI system might find that the symbolic vocabulary is missing relevant concepts.
The system should be capable of identifying such vocabulary incompleteness on its own or at least with the help of the human interacting with the system.

Once the vocabulary incompleteness is identified the next challenge is to work with the human to identify and learn the missing vocabulary items.
In many cases, the system could provide additional assistance to make the symbol acquisition more directed for the problem at hand.
For example, if the system is trying to find more concepts to approximate a certain model component, it could use its knowledge to provide additional information to the human that could help them provide more relevant concepts.
This could, for instance, take the form of the AI agent using low-level explanations, such as saliency maps, to highlight parts of the state or agent behavior relevant to the model component in question.

Note that the problem of detecting vocabulary incompleteness could be particularly tricky in cases where the system may be using the symbols purely as a way to analyze human's non-symbolic input. 
In such cases, the system could incorrectly adopt a hypothesis that is expressed solely in terms of the previously specified concepts, instead of realizing that it requires a new concept to identify the correct human input. One way to avoid such biases may be to ensure that the system always maintains some uncertainty in any hypotheses it learns from the human input \cite{hadfield2017off}.

In the case of super-human AIs, there may be an additional challenge that the human's vocabulary itself isn't sufficient to create a helpful explanation. In such cases, we may need to make use of strategies from intelligent tutoring systems (ITS) \cite{anderson1985intelligent} to enable the AI systems to teach concepts to humans. {An interesting first step taken in this direction is the analysis performed by \citet{mcgrath2021acquisition}, in analyzing the concepts learned by AlphaZero.}


{Note that the process of adding new concepts to the system's vocabulary, teaching new vocabulary items to the human, or even potentially correcting any inconsistency in concept grounding between the two can be seen as part of a {\em vocabulary reconciliation}. This means that the introduction of the symbolic middle layer introduces a whole new dimension to the previously studied model reconciliation process \cite{sreedharan2021foundations}, one where the agent also has to ensure that the shared vocabulary is rich enough to communicate the required model information.
This goes beyond just allowing for noisy groundings.
} 
\section{Case Studies}
In this section, we aim to illustrate some concrete solutions to the challenges of building and maintaining symbolic interfaces with the help of recent work from our research group. 
\subsection{Providing Explanation to Humans}
\label{sec-blackbox}

\begin{figure}[h]
\centering
\includegraphics[scale=0.25]{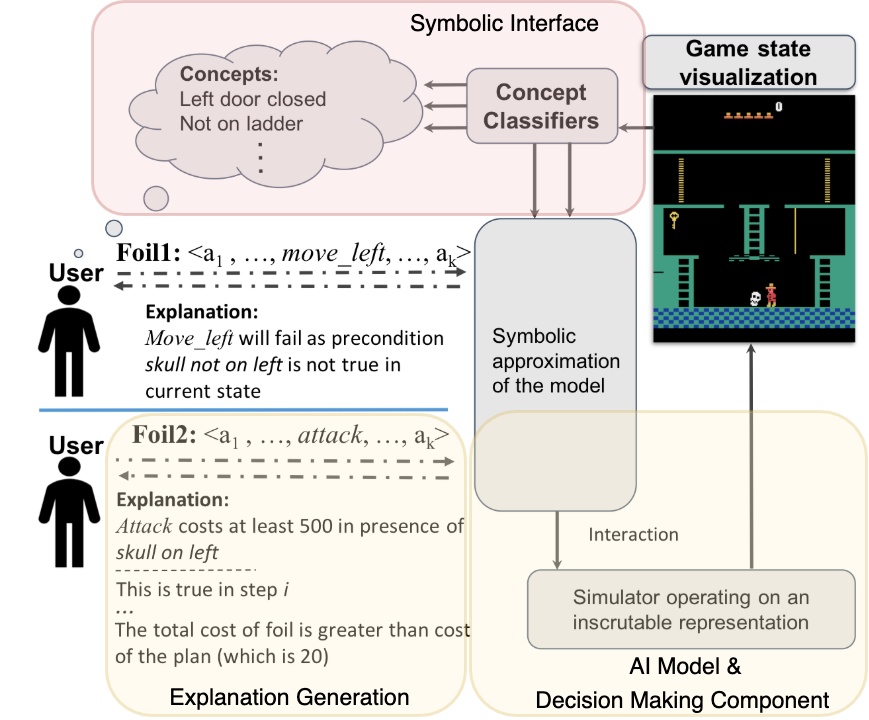}
\caption{Generating symbolic explanations for sequential decisions made over inscrutable internal representations and reasoning (c.f. \cite{sreedharan2020bridging})}

\label{blackboxoverlay}
\end{figure}

We will start with the Blackbox project \cite{sreedharan2020bridging}, which addresses the problem of providing symbolic explanations for sequential decisions made by an AI system over its won inscrutable internal representations and reasoning
{In this work, given an explanatory query, usually, a contrastive one that asks why the current plan was selected over another that the user expected, the system tries to generate an explanation in terms the user understands.} 
The Decision Making Component makes its decisions based on an AI Model that is opaque to the end-user (say a learned model or a simulator).
For a given explanatory query, the work tries to construct parts of a symbolic model (learned from samples generated from the opaque model), particularly missing preconditions and abstract cost function, expressed in human-understandable concepts. This model is then used to provide specific explanations via the Explanation Generation module. In regards to concept sets, the system assumes access to a set of user-specified propositional concepts, along with their corresponding classifiers. These classifiers are grounded based on positive and negative examples for each concept hence establishing a symbolic interface as in Figure \ref{blackboxoverlay}.
The work captures the uncertainty regarding the grounding by using the classification accuracy of the system. Additionally, the system associates a level of uncertainty to each learned symbolic model-component, that not only captures any uncertainty related to grounding but also the fact that the system may have used too few samples to identify the correct model component. Finally, the system can detect that the original vocabulary set may be incomplete if the algorithm is unable to find an explanation for the user query.
\subsection{Interpreting Human Advice}
\label{expand-sec}

\begin{figure}[h]
\centering
\includegraphics[scale=0.16]{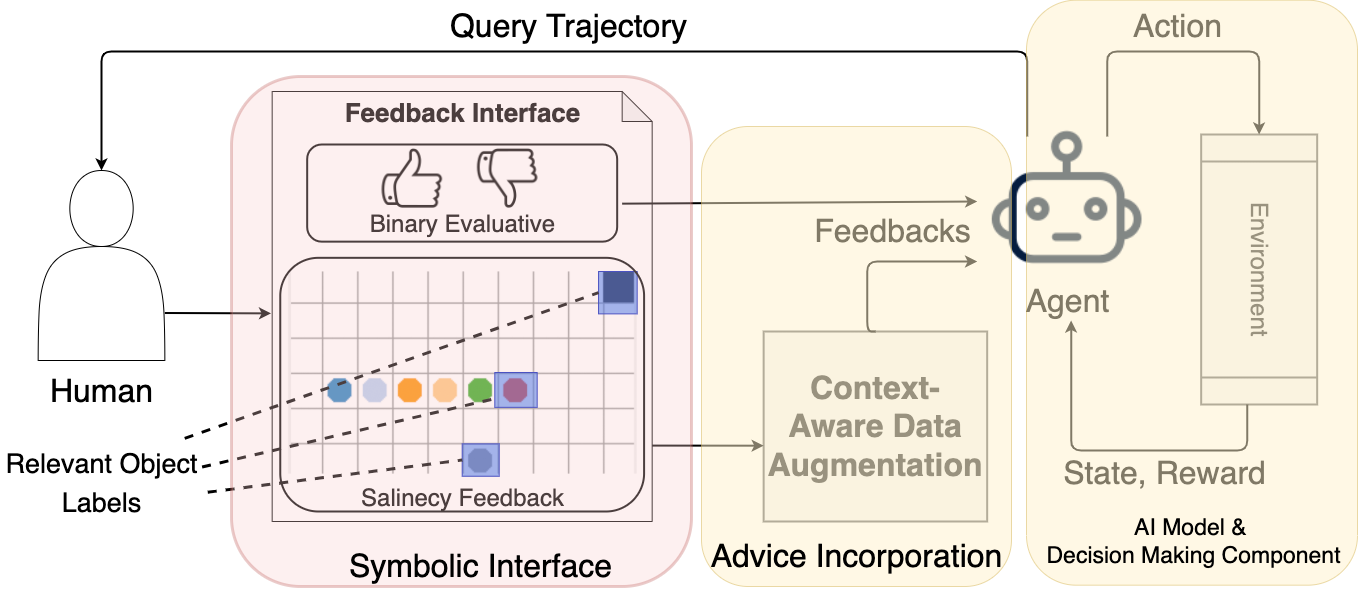}
\caption{Enabling humans to provide quasi-symbolic advice for deep RL systems (c.f. \cite{guan2021widening})}
\label{expandoverlay}
\end{figure}

\begin{figure}[h]
\centering
\includegraphics[scale=0.4]{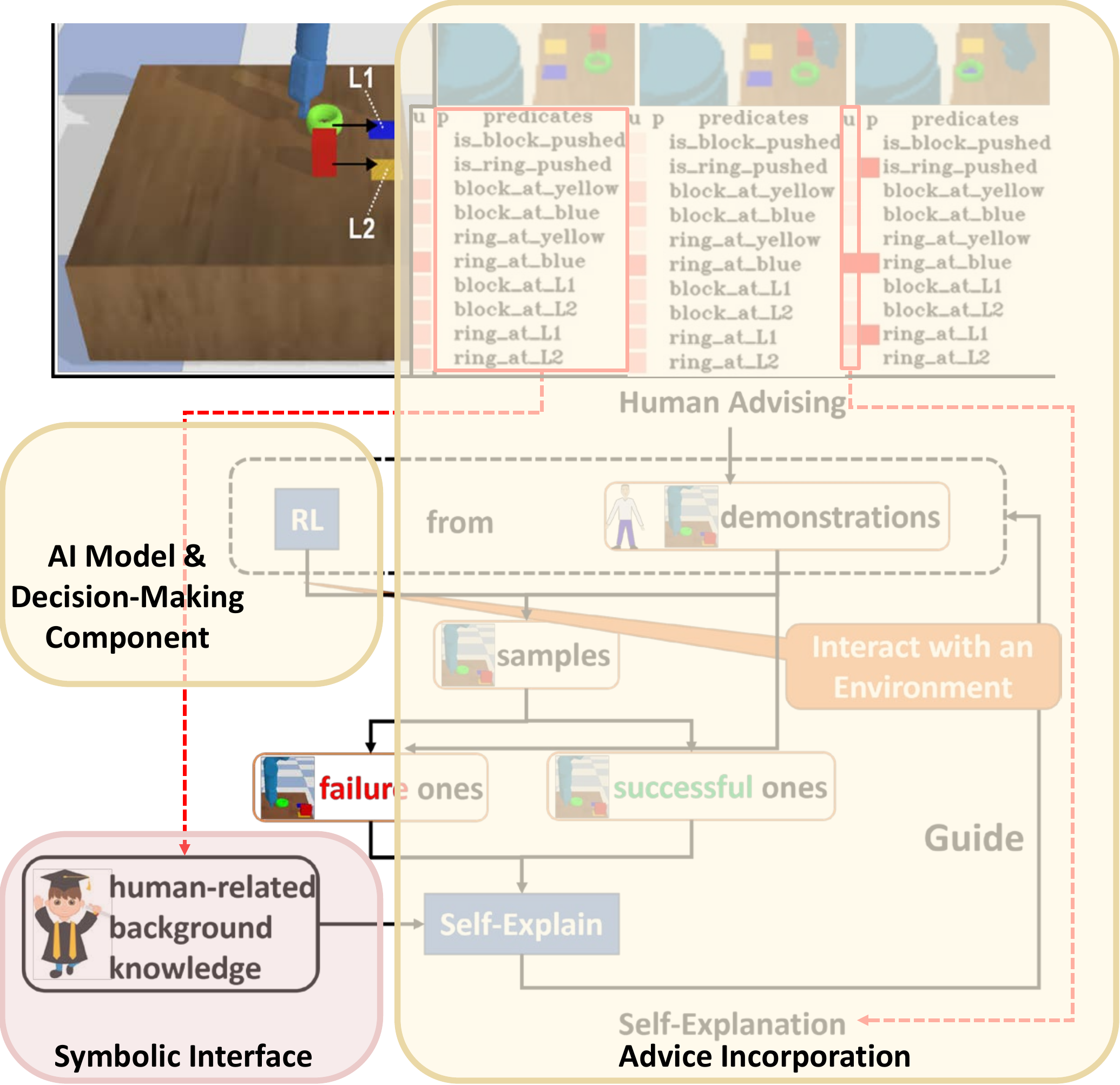}
\caption{Interpreting human demonstrations in the context of the symbolic interface (c.f. \cite{yantian-self-expl}}
\label{fig:serlfd}
\end{figure}

An illustration of the use of  symbolic interfaces to more effectively interpret human advice is provided by  EXPAND system \cite{guan2021widening}, which tries to utilize human binary evaluative feedback and visual explanation to accelerate Human-in-the-Loop Deep Reinforcement Learning. The visual advice is given as saliency regions associated with the action taken. Such feedback could be expensive to collect, as well as unintuitive to specify, especially when the human has to provide such annotations for each query. To make the feedback collection process more efficient and effortless for the human expert, EXPAND leverages an object-oriented interface to convert the labels of relevant objects into corresponding saliency regions in image observations via off-the-shelf object detectors (which effectively ``ground" human object symbols into the image STST and enable a symbolic interface with the human in the loop as shown in Figure \ref{expandoverlay}). 
The human feedback is thus interpreted, as part of Advice-Incorporation, with the assumption that it refers to objects that are relevant from human's point of view. This despite the fact that the Deep RL part of EXPAND is operating over pixel space, and constructing its own internal representations. 
 %
 Object-oriented symbolic interfaces like these have also been used in other previous works to allow humans to provide informative and generalizable object-focused advices in an effortless way \cite{thomaz2006reinforcement, krening2016learning, guan2021widening}.




Another related recent project SERLfD \cite{yantian-self-expl} leverages the symbolic interface to better interpret the human advice. Here the aim is to reduce the ambiguity in human demonstrations of robotic tasks to improve the efficiency of reinforcement learning from demonstrations (RLfD), as illustrated in Fig. \ref{fig:serlfd}. The system assumes that the (continuous) demonstration provided by the human is guided by their own interest in highlighting specific symbolic goals and way points. It learns to interpret the relative importance of these symbols and use that to disambiguate the demonstration (a process that can be viewed as the AI system trying to ``explain'' the demonstration to itself in terms of symbols that are viewed to be critical for the human demonstrator). 

\section{Concluding Remarks}

In this paper, we advocated
an ambitious research program to broaden the basis for
human-AI interaction by proposing that AI systems support a symbolic interface
-- independent of whether their internal operations themselves are done in human-interpretable symbolic means. In the near term, such symbolic interfaces can start in a modest fashion facilitating explanations and advice taking--as illustrated by the case studies we discussed in the previous section. In the long term, more ambitious symbolic interfaces might involve essentially providing a {\em symbolic persona} for an AI agent, regardless of their own internal inscrutable representations and reasoning. Such an endeavor may well involve leveraging all the progress in ``symbolic AI." While all this might seem a little far fetched given the current state of AI, it is worth considering that we humans ourselves may have evolutionarily developed such symbolic persona to communicate with each other. It is only natural for AI systems to follow a similar evolution to coexist with us.\footnote{One intriguing philosophical question is whether the distinction between {\em internal symbolic reasoning} vs. {\em external symbolic interface} will disappear when the machines develop symbolic persona. After all, philosophers have long argued that symbolic language, that we developed essentially to communicate with each other, has gone on to mold our very thinking (e.g. Wittgenstein's (\citeyear{wittgenstein-tractatus}) dictum {\em ``The limits of my language mean the limits of my world''}). The capacity differences between humans and machines may still let machines to not be constrained the way humans are; but this remains to be seen. A separate, and even more longer-term/far fetched consideration is on the role/status of symbolic languages in a distant future where Neuralink-style technologies reduce the need for languages for communication; something perhaps safely beyond scope even for a {\em blue sky} paper.}




\medskip
\noindent 
{\bf Acknowledgments:} This research is supported in part by ONR grants N00014- 16-1-2892, N00014-18-1- 2442, N00014-18-1-2840, N00014-9-1-2119, AFOSR grant FA9550-18-1-0067, DARPA SAIL-ON grant W911NF19-2-0006 and a JP Morgan AI Faculty Research grant. We thank  Been Kim, Dan Weld, Matthias Scheutz, David Smith, the AAAI reviewers, as well as the members of the Yochan research group for helpful discussions and feedback. 

\bibliography{bib}

\begin{thebibliography}{30}
\providecommand{\natexlab}[1]{#1}

\bibitem[{Anderson, Boyle, and Reiser(1985)}]{anderson1985intelligent}
Anderson, J.~R.; Boyle, C.~F.; and Reiser, B.~J. 1985.
\newblock Intelligent tutoring systems.
\newblock \emph{Science}, 228(4698): 456--462.

\bibitem[{Bonet and Geffner(2019)}]{bonet2019learning}
Bonet, B.; and Geffner, H. 2019.
\newblock Learning first-order symbolic representations for planning from the
  structure of the state space.
\newblock In \emph{ECAI}.

\bibitem[{Chakraborti et~al.(2017)Chakraborti, Sreedharan, Zhang, and
  Kambhampati}]{explain}
Chakraborti, T.; Sreedharan, S.; Zhang, Y.; and Kambhampati, S. 2017.
\newblock Plan Explanations as Model Reconciliation: Moving Beyond Explanation
  as Soliloquy.
\newblock In \emph{IJCAI}, 156--163.

\bibitem[{Christiano et~al.(2017)Christiano, Leike, Brown, Martic, Legg, and
  Amodei}]{christiano-preferences}
Christiano, P.~F.; Leike, J.; Brown, T.~B.; Martic, M.; Legg, S.; and Amodei,
  D. 2017.
\newblock Deep Reinforcement Learning from Human Preferences.
\newblock In \emph{30th Annual Conference on Neural Information Processing
  Systems}, 4299--4307.

\bibitem[{De~Giacomo et~al.(2019)De~Giacomo, Iocchi, Favorito, and
  Patrizi}]{de2019foundations}
De~Giacomo, G.; Iocchi, L.; Favorito, M.; and Patrizi, F. 2019.
\newblock Foundations for restraining bolts: Reinforcement learning with
  LTLf/LDLf restraining specifications.
\newblock In \emph{ICAPS}.

\bibitem[{De~Raedt et~al.(2019)De~Raedt, Manhaeve, Dumancic, Demeester, and
  Kimmig}]{deraedt-neuro-symbolic}
De~Raedt, L.; Manhaeve, R.; Dumancic, S.; Demeester, T.; and Kimmig, A. 2019.
\newblock Neuro-symbolic= neural+ logical+ probabilistic.
\newblock In \emph{NeSy'19@ IJCAI, the 14th International Workshop on
  Neural-Symbolic Learning and Reasoning}.

\bibitem[{Garcez et~al.(2019)Garcez, Gori, Lamb, Serafini, Spranger, and
  Tran}]{lamb-neuro-symbolic}
Garcez, A.~d.; Gori, M.; Lamb, L.~C.; Serafini, L.; Spranger, M.; and Tran,
  S.~N. 2019.
\newblock Neural-symbolic computing: An effective methodology for principled
  integration of machine learning and reasoning.
\newblock \emph{arXiv preprint arXiv:1905.06088}.

\bibitem[{Ghorbani et~al.(2019)Ghorbani, Wexler, Zou, and
  Kim}]{ghorbani2019towards}
Ghorbani, A.; Wexler, J.; Zou, J.~Y.; and Kim, B. 2019.
\newblock Towards automatic concept-based explanations.
\newblock In \emph{NeurIPS}, 9273--9282.

\bibitem[{Greydanus et~al.(2018)Greydanus, Koul, Dodge, and
  Fern}]{greydanus2018visualizing}
Greydanus, S.; Koul, A.; Dodge, J.; and Fern, A. 2018.
\newblock {Visualizing and Understanding Atari Agents}.
\newblock In \emph{ICML}.

\bibitem[{Guan et~al.(2021)Guan, Verma, Guo, Zhang, and
  Kambhampati}]{guan2021widening}
Guan, L.; Verma, M.; Guo, S.; Zhang, R.; and Kambhampati, S. 2021.
\newblock Widening the Pipeline in Human-Guided Reinforcement Learning with
  Explanation and Context-Aware Data Augmentation.
\newblock In \emph{Thirty-Fifth Conference on Neural Information Processing
  Systems (Spotlight)}.

\bibitem[{Hadfield-Menell et~al.(2017)Hadfield-Menell, Dragan, Abbeel, and
  Russell}]{hadfield2017off}
Hadfield-Menell, D.; Dragan, A.; Abbeel, P.; and Russell, S. 2017.
\newblock The off-switch game.
\newblock In \emph{3rd International Workshops on AI, Ethics and Society at the
  Thirty-First AAAI Conference on Artificial Intelligence}.

\bibitem[{Icarte et~al.(2018)Icarte, Klassen, Valenzano, and
  McIlraith}]{icarte2018using}
Icarte, R.~T.; Klassen, T.; Valenzano, R.; and McIlraith, S. 2018.
\newblock Using reward machines for high-level task specification and
  decomposition in reinforcement learning.
\newblock In \emph{International Conference on Machine Learning}, 2107--2116.
  PMLR.

\bibitem[{Kahneman(2011)}]{kahneman2011thinking}
Kahneman, D. 2011.
\newblock \emph{Thinking, fast and slow}.
\newblock Macmillan.

\bibitem[{Kambhampati(2020)}]{rao-aimag-haai}
Kambhampati, S. 2020.
\newblock Challenges of Human-Aware {AI} Systems {AAAI} Presidential Address.
\newblock \emph{{AI} Mag.}, 41(3): 3--17.

\bibitem[{Kambhampati(2021)}]{polanyi-cacm}
Kambhampati, S. 2021.
\newblock Polanyi's revenge and AI's new romance with tacit knowledge.
\newblock \emph{Commun. {ACM}}, 64(2): 31--32.

\bibitem[{Konidaris, Kaelbling, and Lozano-Perez(2018)}]{konidaris2018skills}
Konidaris, G.; Kaelbling, L.~P.; and Lozano-Perez, T. 2018.
\newblock From skills to symbols: Learning symbolic representations for
  abstract high-level planning.
\newblock \emph{Journal of Artificial Intelligence Research}, 61: 215--289.

\bibitem[{Krening et~al.(2016)Krening, Harrison, Feigh, Isbell, Riedl, and
  Thomaz}]{krening2016learning}
Krening, S.; Harrison, B.; Feigh, K.~M.; Isbell, C.~L.; Riedl, M.; and Thomaz,
  A. 2016.
\newblock Learning from explanations using sentiment and advice in RL.
\newblock \emph{IEEE Transactions on Cognitive and Developmental Systems},
  9(1): 44--55.

\bibitem[{Krishna et~al.(2017)Krishna, Zhu, Groth, Johnson, Hata, Kravitz,
  Chen, Kalantidis, Li, Shamma et~al.}]{krishna2017visual}
Krishna, R.; Zhu, Y.; Groth, O.; Johnson, J.; Hata, K.; Kravitz, J.; Chen, S.;
  Kalantidis, Y.; Li, L.-J.; Shamma, D.~A.; et~al. 2017.
\newblock Visual genome: Connecting language and vision using crowdsourced
  dense image annotations.
\newblock \emph{International Journal of Computer Vision}, 123(1): 32--73.

\bibitem[{McCarthy(1959)}]{mccarthy1959programs}
McCarthy, J. 1959.
\newblock Programs with Common Sense, proceedings of the Teddington Conference
  on the Mechanization of Thought Processes, 75-91.

\bibitem[{McDermott et~al.(1998)McDermott, Ghallab, Howe, Knoblock, Ram,
  Veloso, Weld, and Wilkins}]{mcdermott1998pddl}
McDermott, D.; Ghallab, M.; Howe, A.; Knoblock, C.; Ram, A.; Veloso, M.; Weld,
  D.; and Wilkins, D. 1998.
\newblock PDDL-the planning domain definition language.

\bibitem[{McGrath et~al.(2021)McGrath, Kapishnikov, Toma{\v{s}}ev, Pearce,
  Hassabis, Kim, Paquet, and Kramnik}]{mcgrath2021acquisition}
McGrath, T.; Kapishnikov, A.; Toma{\v{s}}ev, N.; Pearce, A.; Hassabis, D.; Kim,
  B.; Paquet, U.; and Kramnik, V. 2021.
\newblock Acquisition of Chess Knowledge in AlphaZero.
\newblock \emph{arXiv preprint arXiv:2111.09259}.

\bibitem[{Pearl(2009)}]{pearl2009causality}
Pearl, J. 2009.
\newblock \emph{Causality}.
\newblock Cambridge university press.

\bibitem[{Sreedharan, Chakraborti, and
  Kambhampati(2021)}]{sreedharan2021foundations}
Sreedharan, S.; Chakraborti, T.; and Kambhampati, S. 2021.
\newblock Foundations of explanations as model reconciliation.
\newblock \emph{Artificial Intelligence}, 301: 103558.

\bibitem[{Sreedharan, Kulkarni, and Kambhampati(2021)}]{yochan-xai-book}
Sreedharan, S.; Kulkarni, A.; and Kambhampati, S. 2021.
\newblock \emph{Explainable Human-AI Interaction: A Planning Perspective}.
\newblock Morgan \& Claypool Publishers.

\bibitem[{Sreedharan et~al.(2020)Sreedharan, Soni, Verma, Srivastava, and
  Kambhampati}]{sreedharan2020bridging}
Sreedharan, S.; Soni, U.; Verma, M.; Srivastava, S.; and Kambhampati, S. 2020.
\newblock Bridging the Gap: Providing Post-Hoc Symbolic Explanations for
  Sequential Decision-Making Problems with Inscrutable Representations.
\newblock \emph{ICML-HILL Workshop}.

\bibitem[{Sreedharan, Srivastava, and Kambhampati(2021)}]{sreedharan2021using}
Sreedharan, S.; Srivastava, S.; and Kambhampati, S. 2021.
\newblock Using state abstractions to compute personalized contrastive
  explanations for AI agent behavior.
\newblock \emph{Artificial Intelligence}, 301: 103570.

\bibitem[{Thomaz, Breazeal et~al.(2006)}]{thomaz2006reinforcement}
Thomaz, A.~L.; Breazeal, C.; et~al. 2006.
\newblock Reinforcement learning with human teachers: Evidence of feedback and
  guidance with implications for learning performance.
\newblock In \emph{Proceedings of the AAAI conference on artificial
  intelligence}, 1000--1005. Boston, MA.

\bibitem[{Wittgenstein(1922)}]{wittgenstein-tractatus}
Wittgenstein, L. 1922.
\newblock \emph{Tractatus logico-philosophicus}.
\newblock Harcourt Brace \& Company, Inc.

\bibitem[{Zha, Guan, and Kambhampati(2021)}]{yantian-self-expl}
Zha, Y.; Guan, L.; and Kambhampati, S. 2021.
\newblock Learning from Ambiguous Demonstrations with Self-Explanation Guided
  Reinforcement Learning.
\newblock \emph{AAAI-22 Workshop on Reinforcement Learning in Games and arXiv
  preprint arXiv:2110.05286}.

\bibitem[{Zhang et~al.(2020)Zhang, Walshe, Liu, Guan, Muller, Whritner, Zhang,
  Hayhoe, and Ballard}]{zhang2020atari}
Zhang, R.; Walshe, C.; Liu, Z.; Guan, L.; Muller, K.; Whritner, J.; Zhang, L.;
  Hayhoe, M.; and Ballard, D. 2020.
\newblock Atari-head: Atari human eye-tracking and demonstration dataset.
\newblock In \emph{Proceedings of the AAAI conference on Artificial
  Intelligence}, 04, 6811--6820.

\end{thebibliography}
\newpage
\end{document}